\documentclass[conference]{IEEEtran}
\usepackage{times}
\usepackage[a4paper, margin=19mm, top=25mm, bottom=31mm]{geometry} 
\usepackage{multicol}

\usepackage{enumitem} 

\usepackage{graphics} 
\usepackage{epsfig} 
\usepackage{amsmath} 
\usepackage{amssymb}  

\usepackage{url}
\usepackage[ruled, vlined, linesnumbered]{algorithm2e}
\usepackage{verbatim} 
\usepackage{soul, color}
\usepackage{lmodern}
\usepackage{fancyhdr}
\usepackage[utf8]{inputenc}
\usepackage{fourier} 
\usepackage{array}
\usepackage{makecell}
\usepackage{cite}
\SetNlSty{large}{}{:}
\usepackage{float}     
\usepackage{array}
\usepackage{booktabs}

\usepackage{tikz}        
\usepackage{verbatim}    
\usepackage{lipsum}      
\usepackage{fancyvrb} 

\usepackage{fancyvrb}

\usepackage{tabularx}   

\usetikzlibrary{arrows.meta}  
\usepackage{graphicx}  

\usepackage{float} 
\usepackage{hyperref}
\hypersetup{
    colorlinks=true,
    linkcolor=blue,
    filecolor=blue,
    citecolor=blue,
    urlcolor=blue
}

\ifCLASSINFOpdf
\else
\fi
\begin{document}
%
\title{\fontsize{14}{16}\selectfont \textbf{Optimized Agent Shift Scheduling  Using Multi-Phase Allocation Approach}}


\author{
  \fontsize{12}{12}\selectfont 
  Sanalkumar K, Koushik Dey, Swati Meena \\
  \fontsize{10}{10}\selectfont 
  Sprinklr AI \\
  \texttt{\{sanalkumar.k, koushik.dey, swati.meena\}@sprinklr.com}
}


%


\providecommand{\keywords}[1]
{
  \textbf{\textit{Keywords---}} #1
}

\maketitle

\begin{abstract}
Effective agent shift scheduling is crucial for businesses, especially in the Contact Center as a Service (CCaaS) industry, to ensure seamless operations and fulfill employee needs. Most studies utilizing mathematical model-based solutions approach the problem as a single-step process, often resulting in inefficiencies and high computational demands. In contrast, we present a multi-phase allocation method that addresses scalability and accuracy by dividing the problem into smaller sub-problems of day and shift allocation, which significantly reduces number of computational variables and allows for targeted objective functions, ultimately enhancing both efficiency and accuracy. Each subproblem is modeled as a Integer Programming Problem (IPP), with solutions sequentially feeding into the subsequent subproblem. We then apply the proposed method, using a multi-objective framework, to address the difficulties posed by peak demand scenarios such as holiday rushes, where maintaining service levels is essential despite having limited number of employees. 
\end{abstract}

%
\IEEEpeerreviewmaketitle

\section{INTRODUCTION}
The Contact Center as a Service (CCaaS) industry is rapidly transforming, demanding innovative solutions to manage complex operational challenges, with agent shift scheduling standing out as a critical concern. Agents are the front-line employees responsible for managing customer interactions, responding to inquiries and resolving issues. These interactions are organized into shifts, specific periods during which agents are scheduled to work. Agent shift scheduling involves strategically assigning shifts to agents to balance organizational needs with employee satisfaction. This process ensures that enough skilled agents are present during peak demand times, helping to maintain high service quality essential for customer satisfaction while minimizing unnecessary labor costs. \\

Over the years, the number of interactions between customers and companies has grown significantly with the new channels like emails, chats and social media.  Consequently, traditional call centers have evolved into expansive multi-channel contact centers that seamlessly connect businesses with their customers. This evolution has led to a consistent rise in the demand for workforce numbers and diverse skill sets in contact centers \cite{manyika2017}. However agents are considered a costly resource due to the extensive training and continuous development necessary for them to deliver high-quality customer service. Therefore, efficient shift scheduling is crucial for maximizing the effectiveness of these valuable resources and for ensuring service standards are met. Inefficient agent shift allocation can lead to under-staffing, where too few agents are available, causing long wait times and customer frustration. Over-staffing occurs when too many agents are scheduled, leading to unnecessary labor costs. Both scenarios can significantly impact a business's bottom line and its reputation among customers and employees alike. A standard goal in call center operations is to achieve specific service levels, like answering X\% of calls within Y seconds as outlined in the \textbf{Service Level Agreements (SLAs)}. That's why reducing costs isn't straightforward; maintaining SLAs is essential for high-quality service, as customer experience directly influences profitability\cite{Lywood2009rf}.\\

Agent shift scheduling is crucial in sectors like healthcare \cite{guerriero2022}, retail \cite{esteban2020}, and customer service centers \cite{koole2023, MATTIA201725}, each with unique demands. As scheduling complexity grew, metaheuristic methods emerged to overcome the limitations of traditional models, using advanced techniques to explore complex problem spaces for solutions. Popular metaheuristics include, Genetic Algorithms \cite{saraswati2021}, Variable Neighbourhood Search (VNS) \cite{kenneth2016, kenneth2019}, Simulated Annealing \cite{KLETZANDER2020104794} and many others. However, they often have high time complexity and require multiple runs for stable results due to randomness. \\

In real-world agent shift scheduling, organizations must balance multiple, often conflicting objectives—such as maximizing operational efficiency, minimizing costs etc. The complexity of balancing multiple objectives has led to multi-objective optimization frameworks, often combining objectives with weighted empirical assignments \cite{ryu2023, mahalakshmi2020}. However, these approaches frequently struggle to manage these competing objectives, as they require compromises that can result in suboptimal or impractical solutions and the effectiveness of the generated schedule can be compromised if the weights are altered, often making time consuming when solved using a single optimization model. We found that this challenge becomes even more pronounced during peak periods, when a surge in customer requests coincides with limited agent availability, making it difficult to maintain both service quality and operational efficiency.
\begin{table*}[!t]
\centering
\fontsize{10}{10}\selectfont T\fontsize{8}{8}\selectfont{\uppercase{able I}} \\
\vspace{0.1cm}
\fontsize{8}{8}\selectfont N\fontsize{7}{7}\selectfont {\uppercase{OTATIONS WITH DESCRIPTIONS AND RANGES}}
\vspace{0.15cm}
\renewcommand{\arraystretch}{1.2} 
\setlength{\tabcolsep}{6pt} 

\begin{tabular}{|p{0.15\textwidth}|p{0.5\textwidth}|p{0.35\textwidth}|}
\hline
\textbf{Set Notation} & \textbf{Description} & \textbf{Set Elements} \\ \hline
$A$ & Set of all agents. & $a \in A$; $a$ is an independent Customer Center agent.\\
$D$ & Set of all days in the scheduling period. & $d \in D$; $d$ is an individual date. \\
$T$ & Set of time intervals in a day, dividing 24 hours into equal segments. & $t \in T$; $t$ denotes a different hour of the day. \\
$S$ & Set of all available shifts in the schedule. & $s \in S$, where $s = (t, t + \text{duration of shift})$; $t \in T$ \\
$W$ & Set of weeks in the scheduling period. & $w \in W$; note that $w$ is a subset of $D$ \\
\hline
\textbf{Notation} & \textbf{Description} & \textbf{Range} \\ \hline
$R_{D}(d)$ & Total number of staff required on day $d$. & $R_{D}(d) \in \mathbb{Z}^{+}, \, \forall d \in D$ \\
$R_{DT}(d, t)$ & Number of agents required during interval $t$ on day $d$. & $R_{DT}(d, t) \in \mathbb{Z}^{+}, \, \forall d \in D, \, \forall t \in T$ \\
$P_{D}(d)$ & Number of agents assigned by the schedule on day $d$. & $P_{D}(d) \in \mathbb{Z}^{+}, \, \forall d \in D$ \\
$P_{DT}(d, t)$ & Number of agents assigned by the schedule to an interval $t$ on day $d$. & $P_{DT}(d, t) \in \mathbb{Z}^{+}, \, \forall d \in D, \, \forall t \in T$ \\
$B_{AD}(a, d)$ & $B_{AD}(a, d) = 1$ if agent $a$ works on day $d$, else $0$. & $B_{AD}(a, d) \in \{0, 1\}, \, \forall a \in A, \, \forall d \in D$ \\
$B_{ADS}(a, d, s)$ & $B_{ADS}(a,d,s) = 1$ if agent $a$ works shift $s$ on day $d$, else $0$. & $B_{ADS}(a,d,s) \in \{0, 1\}, \, \forall a \in A, \, \forall d \in D, \, \forall s \in S$ \\
$C_{ADS}(a, d, s)$ & Cost of scheduling agent $a$ in shift $s$ on day $d$. & $C_{ADS}(a,d,s) \geq 0$ \\
$U_{D}(d)$ & Amount of under/over staffing on day $d$, integer variable. & $U_{D}(d) = R_{D}(d) - P_{D}(d), \, \forall d \in D$ \\
$U_{DT}(d, t)$ & Amount of under/over staffing during interval $t$ on day $d$, integer variable. & $U_{DT}(d, t) = R_{DT}(d, t) - P_{DT}(d, t), \, \forall d \in D, \, \forall t \in T$ \\
$V_{D}(d)$ & Agent day allocation penalty variable for day $d$, integer variable. & $V_{D}(d) \in \mathbb{Z}, \, \forall d \in D$ \\
$I_{ST}(s, t)$ & $I_{ST}(s,t) = 1$ if shift $s$ covers interval $t$, else $0$. & $I_{ST}(s,t) \in \{0, 1\}, \, \forall s \in S, \, \forall t \in T$ \\
\hline
$K$ & Day allocation penalty factor. & $K \in \mathbb{Z}^{+}$\\
$\epsilon$ & Small constant used for numerical stability. & $\epsilon \geq 0$ \\
\hline
\end{tabular}
\label{table:full_width}
\end{table*}

To address these gaps, our paper makes the following key contributions: \\
\begin{itemize}
\item \textbf {\textit{Multi-Phase Allocation Approach}}:
We propose a structured method that divides the problem into agent day and shift allocation, each formulated as an Integer Programming Problem (IPP). This approach enables independent optimization at each stage, resulting in more practical and effective scheduling solutions.
\item \textbf {\textit{Peak Season Optimization with Limited Staffing}}: We introduce balanced objectives, specifically designed for peak season scenarios, ensuring optimal service levels with limited agents.\\
\end{itemize}
\section{METHODOLOGY}
\vspace{0.1cm}
Agent shift scheduling involves strategically assigning the right number of agents (\(a\)) across different days (\(d\)) and shifts (\(s\)) to ensure staffing levels closely match fluctuating customer demand. The primary goal is to have enough agents available at all times to meet customer needs efficiently, without wasting resources. Going forward, all notations used in the work can be referred from 
\textit{Table I}. Any additional notations will be defined as needed in the relevant sections. 
\vspace{0.15cm}

For simplicity, we consider two key business constraints in our scheduling model: 
\begin{itemize}
    \item \textit{Each agent will be working 5 days in a week.}
    \item \textit{Each agent is assigned to maximum one shift in a working day.}
\end{itemize}
The core objective in agent shift scheduling is to minimize total scheduling costs. This can be mathematically expressed as:
\vspace{0.2cm}
\begin{align}
\text{Minimize } \sum_{a \in A} \sum_{d \in D} \sum_{s \in S}  B_{ADS}(a,d,s) \cdot C_{ADS}(a, d, s)
\\ \nonumber
\end{align}
Although above minimizes the total cost, in real life scenarios the total number of agents are generally fixed and we need to assign those agents to a schedule such that 
\begin{itemize}
    \item \textit{We minimize the under-staffing (difference of required and assigned agents ) in any interval.}
    \item \textit{We also do not overstaff (-ve of under-staffing) in any interval.}
\end{itemize}

Taking these considerations into account, the objective can be reformulated as:
\begin{align}
\text{Minimize:   } & \ \sum_{d \in D} \sum_{t \in T} \left[ R_{DT}(d,t) - P_{DT}(d,t) \right]^2 \\
\text{Minimize:  } & \ \sum_{d \in D} \sum_{t \in T} \left(U_{DT}(d, t)\right)^{2}
\end{align}
This formulation emphasizes aligning staffing closely with demand, thereby improving service quality and operational efficiency. Here \( R_{DT}(d,t) \) is precomputed using the \textit{Erlang-C} formula based on call volume, Average Handling Time (AHT) and SLA thresholds. In our study, we have used a 5-minute AHT and an 80/20 SLA.
\section*{Multi-Phase Allocation Approach }
\vspace{0.15cm}
We now employ the multi-phase allocation approach by breaking the problem into sub-problems of agent day and shift allocation so that specific business-driven constraints and objectives can be applied in each phase. This strategic approach addresses the inherent complexity and computational challenges of optimizing both day and shift allocations simultaneously. 

\subsection*{Agent Day Allocation}
\vspace{0.1cm}
We first determine the specific days each agent will work using day allocation-specific constraints. Agent day allocation should be based on the requirement for each day, ensuring more agents are scheduled on high-demand days and fewer on lighter days.\\ \\ 
\textbf{Constraints:}
\begin{itemize}
    \item \textit{Each agent will be working 5 days in a week.} 
    \begin{align}
        & \sum_{d \in \text{w}} B_{AD}(a, d) = 5, \quad \forall a \in A, \forall w \in W
    \end{align}
    \item \textit{Referring to Table I, we can define \( U_{D}(d) \) as:}
\begin{align}
    &U_{D}(d) = R_{D}(d) - P_{D}(d), \quad \forall d \in D
\end{align}
\begin{align}
    &R_{D}(d) := \max_{t} R_{DT}(d, t); \quad P_{D}(d) = \left( \sum_{a \in A} B_{AD}(a, d) \right) \nonumber
\end{align}
\end{itemize}
\textbf{Objective:} 
\begin{itemize}
\item \textit {Align agent allocation with daily demand to minimize under-staffing and over-staffing on any day.}
\begin{align}
\text{Minimize:} \quad \sum_{d \in D} (U_{D}(d))^2 
\end{align}
\end{itemize}
\subsection*{Agent Shift Allocation}
Now, let's determine which shift each agent will be assigned to on their respective working days based on shift allocation-specific constraints. Here, we will not create variables for all combinations of agent, day, and shift. Instead, we use the agent-day allocation results to create variables only for the pairs where agents are scheduled to work. We have the number of agents assigned by above phase is given by $n_d$ for a given day $d$.\\ 
\textbf{Constraints:}
\begin{itemize}
\item \textit{An agent will be working single shift in a day.}
\begin{align}
& \sum_{s \in \text{S}} B_{ADS}(a, d, s) = 1, \quad \forall a \in A, \forall d \in D
\end{align}
\item \textit{Number of agents working in a day is precomputed from above phase.}
\begin{align}
& \sum_{a \in A} \sum_{s \in S} B_{ADS}(a, d, s) = n_d, \quad \forall d \in D
\end{align}
\item \textit{Referring to Table I, we can define \( U_{DT}(d,t) \) as:}
\begin{align}
    &U_{DT}(d,t) = R_{DT}(d,t) - P_{DT}(d,t), \quad \forall d \in D, \forall t \in T \\
    &\text{where} \quad P_{DT}(d,t) = \left( \sum_{a \in A} \sum_{s \in S} B_{ADS}(a,d,s) \right) \cdot I(s,t) \nonumber
\end{align}   
\end{itemize}
\textbf{Objective:} $\rightarrow$ \textit{Equation(3)}.
\vspace{0.1cm}
\section*{Single-Phase Optimization Approach}
\vspace{0.15cm}
For comparison, we also generated schedules using a single-phase method. This approach models the entire agent shift scheduling problem as a single integer programming problem. The results from this method serve as a baseline for evaluating our proposed approach. Let’s formulate the IPP by defining the relevant business constraints and objective.\\ \\
\textbf{Constraints:}
\begin{itemize}
\item \textit{Each agent will be working 5 days in a week.} 
\begin{align}
& \sum_{s \in S} \sum_{d \in \text{w}} B_{ADS}(a, d, s) = 5, \quad \forall a \in A, \, \forall w \in W
\end{align}

\item \textit{An agent will be working single shift in a day.}
\begin{align}
& \sum_{s \in \text{S}} B_{ADS}(a, d, s) =
\begin{cases} 
1, & \text{if agent } a \text{ works on day } d \\
0, & \text{otherwise}
\end{cases}
\end{align}
\item \textit{Referring to Table I, we can define \( U_{DT}(d,t) \) as: $\rightarrow$ Equation(9).}
\end{itemize}
\vspace{0.1cm}
\textbf{Objective:} $\rightarrow$ \textit{Equation(3)}.
\vspace{0.15cm}
\section*{Peak Season Optimization with Limited Staffing}
\vspace{0.15cm}
Imagine a bustling call center business facing a holiday rush, where customer inquiries surge beyond the available staff. There’s an urgent need to maintain service levels, but budget constraints limit hiring additional agents. Such challenges aren’t unique, often occurring during peak seasons or unexpected events. In these scenarios, even with limited staff, it's crucial to allocate agents based on the required agent data distribution. We have generated schedules for 70 agents over one week using the agent day allocation phase, as outlined in (4)(5)(6). The week starts with a high demand, averaging 225 agent requirements on weekdays and 110 on weekends, both exceeding our total of 70 available agents.

\begin{figure}[h]
    \centering
    \includegraphics[width=0.5\textwidth]{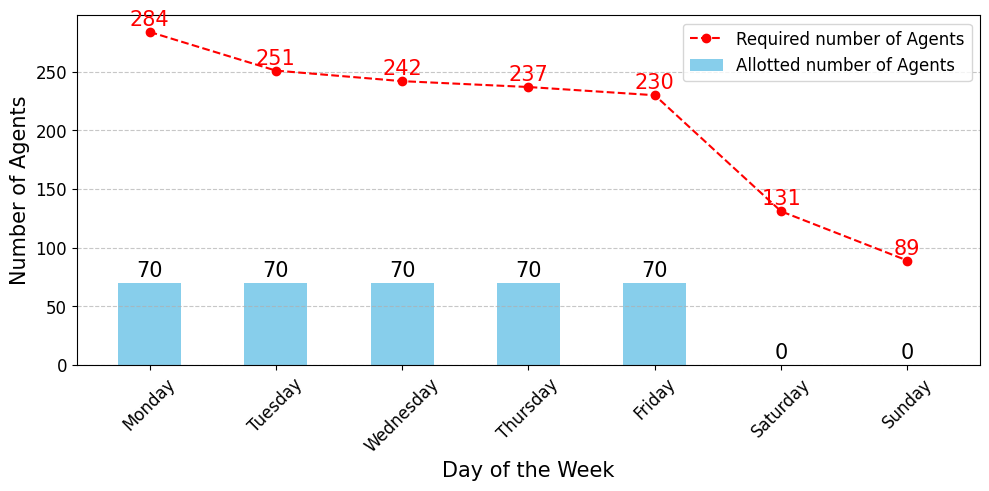}
    \caption{{\fontsize{8}{9.6}\selectfont Required vs Allotted Agents on Each Day of the Week}}
    \label{fig:image_label}
\end{figure}

In analyzing the agent day allocation (\textit{Fig-1}), we observed a significant issue, agent day allocation was concentrated on weekdays, resulting in no staff coverage during weekends and thereby effectively shutting down operations on those days. While this distribution theoretically minimizes the total under-staffing and over-staffing across the schedule, it is impractical from a business operations standpoint where consistent daily staffing is essential.\\

To tackle this issue, we have added a penalty term that imposes a substantial cost when agents are predominantly scheduled on specific days. This encourages a more balanced distribution of staff across all days, ensuring operational continuity. A second consideration is how to evaluate whether one schedule is superior to another. Our objective is to create a schedule that closely aligns with daily agent requirements. For this comparison, we use the \textit{Kullback-Leibler (KL) Divergence} metric, which quantifies the difference between two probability distributions. A lower KL divergence value indicates a closer match, thereby signifying a more optimal schedule.\\
We can define the day allocation penalty as follows:
\begin{align}
    &V_{D}(d) = K \cdot (|A| - P_{D}(d)) \quad \forall d \in D
\end{align}
\textbf{Modified Objective:} 
We employ two objectives here: the first aims to minimize day-level under-staffing and over-staffing, thereby ensuring adequate coverage for each day; the second aims to align the agent day allocation with the required day distribution by minimizing their KL divergence.
\begin{align}
\text{Minimize:} \quad \sum_{d \in D} (U_{D}(d))^2 + (V_{D}(d))^2 
\end{align}
\begin{align}
K^* = \arg \min_{K} D_{KL}(\lambda_{K} \parallel \alpha)
\end{align}
\textbf{Where: }
\vspace{0.1cm}
\begin{itemize}[leftmargin=4em] 
    \item \( \lambda_{K} = \left[ \frac{P_D(d_1)}{\sum_{d \in D} P_D(d)}, \; \frac{P_D(d_2)}{\sum_{d \in D} P_D(d)}, \; \ldots, \; \frac{P_D(d_{|D|})}{\sum_{d \in D} P_D(d)} \right] \)
    \vspace{0.25cm}
    \item \( \alpha = \left[ \frac{R_D(d_1)}{\sum_{d \in D} R_D(d)}, \; \frac{R_D(d_2)}{\sum_{d \in D} R_D(d)}, \; \ldots, \; \frac{R_D(d_{|D|})}{\sum_{d \in D} R_D(d)} \right] \)
    \vspace{0.25cm}
    \item \( D_{KL}(\lambda_{K} \parallel \alpha) = \sum_{d \in D} \lambda_{K} \log \left( \frac{\lambda_{K}}{\alpha + \epsilon} \right) \)
    \vspace{0.25cm}
\end{itemize}
For each integer \(K\) starting from \(0\), we obtain an allocation distribution \(\lambda_{K}\) using (13) and compute the corresponding KL divergence. This iterative process continues for each  \(K\) until the least KL divergence value is identified (14). We observe that the KL loss forms an elbow curve for each integer, and upon reaching its minimum, we cease further iterations for subsequent penalty values.
\begin{table*}[!h]
\centering
\fontsize{10}{10}\selectfont T\fontsize{8}{8}\selectfont{\uppercase{able III} \\
\vspace{0.1cm}
\fontsize{8}{8}\selectfont S\fontsize{7}{7}\selectfont {\uppercase{INGLE-PHASE AND MULTI-PHASE: PERFORMANCE COMPARISON}} 
\vspace{0.15cm}
\renewcommand{\arraystretch}{1.4} 
\setlength{\tabcolsep}{7pt} 

\begin{tabular}{|c|c|c|c|c|c|c|c|c|c|}
\hline
\multicolumn{3}{|c|}{\textbf{Configuration}} & \multicolumn{2}{|c|}{\textbf{Number of Variables}} & \multicolumn{2}{c|}{\textbf{IVDI}} & \multicolumn{2}{c|}{\textbf{DVDI}} \\ 
\hline
\textit{No. of Agents} & \textit{No. of Days} & \textit{Time Limiter} & \textit{Single-Phase} & \textit{Multi-Phase} & \textit{Single-Phase} & \textit{Multi-Phase} & \textit{Single-Phase} & \textit{Multi-Phase} \\ 
\hline
250 & 30 & 10 Min & 113940 & \textbf{91470 ($\downarrow$)} & 80116 & \textbf{12584 ($\downarrow$)} & 2539 & \textbf{691 ($\downarrow$)} \\ 
\hline
250 & 45 & 12 Min & 170910 & \textbf{127105 ($\downarrow$)} & 128792 & \textbf{18186 ($\downarrow$)} & 3816 & \textbf{860 ($\downarrow$)}\\ 
\hline
250 & 60 & 15 Min & 227880 & \textbf{179190 ($\downarrow$)} & 179526 & \textbf{12432 ($\downarrow$)} & 5183 & \textbf{1031 ($\downarrow$)}\\ 
\hline
\end{tabular}
}
\label{table:agent_shift_scheduling}
\end{table*}

\begin{figure*}[!h]
    \centering
     \includegraphics[width=0.3\textwidth]{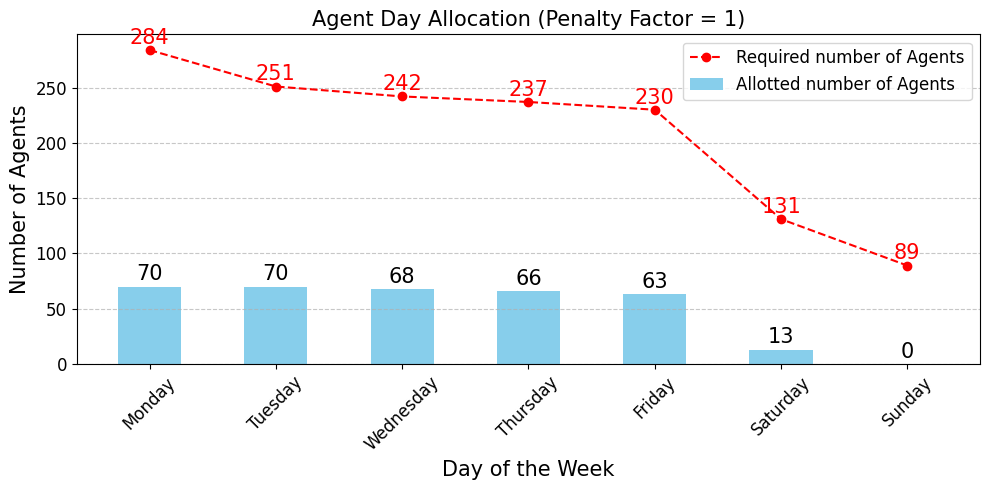}
    \includegraphics[width=0.3\textwidth]{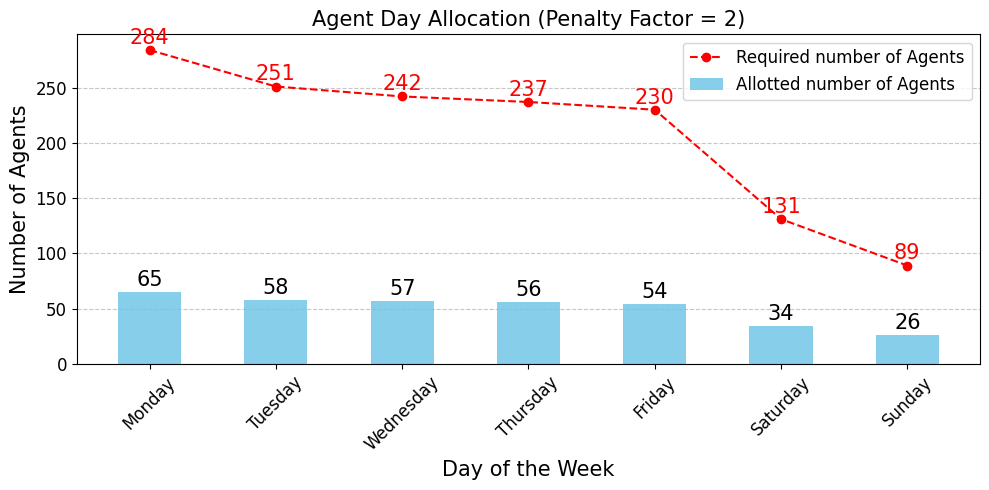}
    \includegraphics[width=0.3\textwidth]{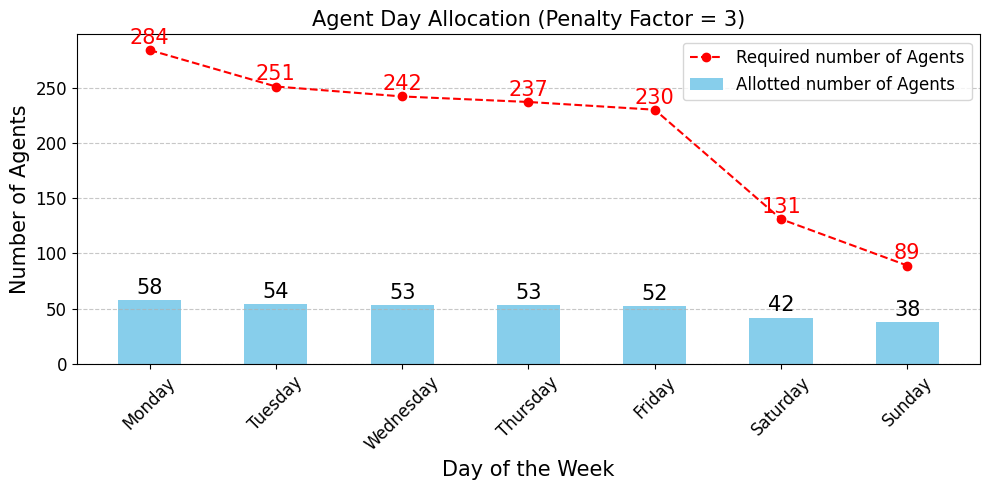}
    \caption{{\fontsize{8}{9.6}\selectfont Penalty Factor vs Agent Day Allocation}}
    \label{fig:combined_images}
\end{figure*}
\section{RESULTS AND DISCUSSION}
\vspace{0.2cm}
This section presents the results from each method, highlighting their effectiveness and comparing their impact on scheduling efficiency. All computations were performed on a device with an M1 processor featuring a 10-core CPU. We have used \textit{Google OR-Tools}  \footnote{\url{https://developers.google.com/optimization/}} as the solver for all computations. 
\section*{Multi-Phase Allocation Approach }
\vspace{0.15cm}
By dividing the scheduling problem into two sub-problems; day allocation and shift allocation, the total number of variables is significantly reduced compared to single-phase approach (\textit{Table II}). The agent day allocation phase determines which days each agent will work. This significantly reduces the variables for agent shift allocation phase by limiting them to only those agent-day pairs selected in the day allocation, rather than considering all possible (agent,day,shift) combinations. Let \( M \) be the number of days selected from the agent day allocation, where \( M < D \).

\begin{table}[H]
\centering
\fontsize{10}{10}\selectfont T\fontsize{8}{8}\selectfont{\uppercase{able II} \\
\vspace{0.1cm}
\fontsize{8}{8}\selectfont S\fontsize{7}{7}\selectfont{\uppercase{ingle-Phase and Multi-Phase: Complexity Analysis}} 
\vspace{0.15cm}
\renewcommand{\arraystretch}{1.4} 
\setlength{\tabcolsep}{7pt} 

\renewcommand{\arraystretch}{1.2} 
\begin{tabular}{|l|c|}
    \hline
    \textbf{Approach} & \textbf{Total Number of Variables Complexity} \\
    \hline
    \textit{Single-Phase Approach} & 
    \( V_1 = O(|A| \times |D| \times |S|) \) \\
    \hline
    \textit{Multi-Phase Approach} & 
    \begin{tabular}{@{}c@{}}
    \( V_2 = O(|A| \times |D|) + O(|A| \times |M| \times |S|) \) \\
    \( = O(|A| \times (|D| + |M| \times |S|)) \)
    \end{tabular} \\
    \hline
\end{tabular}
}
\label{tab:approaches_comparison}
\end{table}
By addressing day and shift allocations separately, distinct objective functions can be utilized for each phase, allowing for more specific and targeted optimization without compromising overall accuracy. To evaluate scheduling performance, we introduce two metrics: the \textit{Day Volume Deviation Index (DVDI)} and the \textit{Interval Volume Deviation Index (IVDI)}, defined as follows:
\begin{align}
\text{DVDI} &= \sum_{d \in D} \left| R_{D}(d) - P_{D}(d) \right|
\end{align}
\begin{align}
\text{IVDI} &= \sum_{d \in D} \sum_{t \in T} \left| R_{DT}(d, t) - P_{DT}(d, t) \right|
\end{align}
We have  generated schedules for a team of 250 agents over 30, 45, and 60 days periods using both single-phase and multi-phase to analyze and compare the results effectively. We've set a solution limiter for each configuration to ensure the process doesn't run excessively long. Both single and multi-phase approaches are given the same computation time. In the single-phase approach, the entire time is dedicated to finding the agent shift allocation solution. Multi-phase approach divides the time between phases. Typically, day allocation requires less time, so we allocate around \textbf{20\%} of the total time for day allocation and the rest for shift allocation. Each configuration was run \textbf{10} times to ensure reliability and the average results are presented in \textit{Table III}.\\

The multi-phase scheduling method demonstrates significant improvements over the single-phase approach. It effectively reduces the number of variables by approximately by \textbf{19-22\%}. Most notably, the IVDI  sees a reduction of \textbf{84-93\%} and DVDI is decreased by \textbf{73-80\%}. These metrics, though excluded from the objective function due to their poor performance in generating diverse distributions, offer an intuitive understanding of schedule quality for an agent supervisor which helps in assessing operational performance. This efficiency stems from separating day and shift allocations, allowing for targeted optimization that addresses specific scheduling needs without compromising accuracy. The single-phase method, on the other hand, handles both tasks simultaneously, leading to conflicting objectives and reduced effectiveness.These improvements may vary based on agent distribution, number of days, agents, and constraints.
\section*{Peak Season Optimization with Limited Staffing}
\vspace{0.15cm}
When the number of agents is significantly lower than actual requirement, and staff requirements are on different scales for each day, solving agent day allocation using (4)(5)(6) will result in allocating more agents to high-demand days, potentially shutting down operations on lower-demand days (\textit{Fig-1}). Along with existing constraints (4)(5), this issue is addressed by incorporating a penalty calculation term (12) and modifying the objective from (3) to those in (13) and (14). Let's analyze how day allocation changes with varying penalty factors. As the penalty factor increases, we impose a greater penalty when agents take day-offs on the same days. \textit{Fig-2} illustrate this trend, showing that a penalty factor of 2 achieves the lowest KL divergence in our case, indicating the most optimal day allocation distribution. This approach is effective because it allows us to focus on day allocation independently and to tweak the objective based on business requirements whereas a single-phase framework must handle multiple conflicting objectives simultaneously, resulting in complex trade-offs that reduce optimization effectiveness. Once the optimal day allocation distribution is achieved, we pass this solution on to tackle the agent shift allocation problem using constraints (7)(8)(9) and objective (3). \\
\section{CONCLUSION}
We proved that multi-phase allocation is an effective methodology for agent shift scheduling in the CCaaS industry. By dividing the problem into sequential day and shift allocation subproblems, the approach significantly reduces computational variables and enables targeted objective functions, thus improving both efficiency and accuracy in daily and interval staffing compared to single-phase method. A key contribution of this work is the adapted day allocation phase tailored for peak seasons with limited staffing. By incorporating a KL divergence–based penalty, the method achieves balanced agent distribution and maintains operational continuity on lower-demand days. Looking ahead, future research can further advance this framework by introducing complex constraints, such as agent preferences, agent proficiency levels, multi-skilled agents and detailed labor laws necessitating dependency on day and shift. Additionally, integrating activity allocation by optimizing breaks within each working shift according to business needs can significantly improve operational efficiency and overall service quality.
\bibliographystyle{IEEEtran}  
\bibliography{references}

@techreport{manyika2017,
  author = {J. Manyika and S. Lund and M. Chui and J. Bughin and J. Woetzel and P. Batra and R. Ko and S. Sanghvi},
  title = {Jobs lost, jobs gained: Workforce transitions in a time of automation},
  year = {2017},
  institution = {McKinsey Global Institute}
}

@article{Lywood2009rf,
  title    = "Customer experience and profitability: An application of the
              empathy rating index ({ERIC}) in {UK} call centres",
  author   = "Lywood, Jamie and Stone, Merlin and Ekinci, Yuksel",
  journal  = "Journal of Database Marketing \& Customer Strategy Management",
  volume   =  16,
  number   =  3,
  pages    = "207--214",
  month    =  sep,
  year     =  2009
}

@article{guerriero2022,
  title    = "Modeling a flexible staff scheduling problem in the Era of
              Covid-19",
  author   = "Guerriero, Francesca and Guido, Rosita",
  journal  = "Optimization Letters",
  volume   =  16,
  number   =  4,
  pages    = "1259--1279",
  month    =  may,
  year     =  2022

}

@article{esteban2020,
title = {Efficient shift scheduling with multiple breaks for full-time employees: A retail industry case},
journal = {Computers and Industrial Engineering},
volume = {150},
pages = {106884},
year = {2020},
issn = {0360-8352},
doi = {https://doi.org/10.1016/j.cie.2020.106884},
author = {Esteban Álvarez and Juan-Carlos Ferrer and Juan Carlos Muñoz and César Augusto Henao}
}

@article{MATTIA201725,
title = {Staffing and scheduling flexible call centers by two-stage robust optimization},
journal = {Omega},
volume = {72},
pages = {25-37},
year = {2017},
issn = {0305-0483},
doi = {https://doi.org/10.1016/j.omega.2016.11.001},
author = {Sara Mattia and Fabrizio Rossi and Mara Servilio and Stefano Smriglio}
}

@article{koole2023,
  author    = {Ger M. Koole and Siqiao Li},
  title     = {A Practice-Oriented Overview of Call Center Workforce Planning},
  journal   = {Stochastic Systems},
  volume    = {13},
  issue    = {4},
  pages     = {479--495},
  year      = {2023},
  doi       = {10.1287/stsy.2021.0008}
}

@article{saraswati2021,
author = {Saraswati, Ni Wayan and Artakusuma, I and Indradewi, I},
year = {2021},
month = {03},
pages = {012014},
title = {Modified genetic algorithm for employee work shifts scheduling optimization},
volume = {1810},
journal = {Journal of Physics: Conference Series},
doi = {10.1088/1742-6596/1810/1/012014}
}

@INPROCEEDINGS{kenneth2016,
  author={Reid, Kenneth N. and Li, Jingpeng and Swan, Jerry and McCormick, Alistair and Owusu, Gilbert},
  booktitle={2016 IEEE Symposium Series on Computational Intelligence (SSCI)}, 
  title={Variable Neighbourhood Search: A case study for a highly-constrained workforce scheduling problem}, 
  year={2016},
  volume={},
  number={},
  pages={1-6},
  doi={10.1109/SSCI.2016.7850087}}

@inproceedings{kenneth2019,
author = {Reid, Kenneth N. and Li, Jingpeng and Brownlee, Alexander and Kern, Mathias and Veerapen, Nadarajen and Swan, Jerry and Owusu, Gilbert},
title = {A hybrid metaheuristic approach to a real world employee scheduling problem},
year = {2019},
isbn = {9781450361118},
publisher = {Association for Computing Machinery},
address = {New York, NY, USA},
doi = {10.1145/3321707.3321769},
booktitle = {Proceedings of the Genetic and Evolutionary Computation Conference},
pages = {1311–1318},
numpages = {8},
location = {Prague, Czech Republic},
series = {GECCO '19}
}

@article{KLETZANDER2020104794,
title = {Solving the general employee scheduling problem},
journal = {Computers and Operations Research},
volume = {113},
pages = {104794},
year = {2020},
issn = {0305-0548},
doi = {https://doi.org/10.1016/j.cor.2019.104794},
author = {Lucas Kletzander and Nysret Musliu}
}

@article{mahalakshmi2020,
  author    = {Mahalakshmi R and Murugesan R and Ranjini K S},
  title     = {Multi Objective Flexible Employee Scheduling Using PSBCSP Algorithm},
  journal   = {International Journal of Emerging Technologies and Innovative Research},
  volume    = {7},
  issue     = {9},
  pages     = {984--993},
  year      = {2020},
  month     = {September},
  issn      = {2349-5162}
}

@Article{ryu2023,
AUTHOR = {Ryu, Hee Jun and Jo, Ye Na and Lee, Won Jun and Cheong, Ji Won and Moon, Boo Yong and Ko, Young Dae},
TITLE = {An Effective Staff Scheduling for Shift Workers in Social Welfare Facilities for the Disabled},
JOURNAL = {Algorithms},
VOLUME = {16},
YEAR = {2023},
NUMBER = {1},
ARTICLE-NUMBER = {41},
ISSN = {1999-4893},
DOI = {10.3390/a16010041}
}
\end{document}